\documentclass[10pt,twocolumn,letterpaper]{article}

\usepackage[pagenumbers]{cvpr} 

\usepackage{graphicx}
\usepackage{amsmath}
\usepackage{amssymb}
\usepackage{booktabs}
\usepackage{wrapfig}
\usepackage{threeparttable}
\usepackage{graphicx}
\usepackage{multicol}
\usepackage{multirow}
\usepackage{pifont}
\usepackage{appendix}

\usepackage[pagebackref,breaklinks,colorlinks]{hyperref}

\usepackage[capitalize]{cleveref}
\crefname{section}{Sec.}{Secs.}
\Crefname{section}{Section}{Sections}
\Crefname{table}{Table}{Tables}
\crefname{table}{Tab.}{Tabs.}

\def\cvprPaperID{7600} 
\def\confName{CVPR}
\def\confYear{2022}

\begin{document}

\title{QAHOI: Query-Based Anchors for Human-Object Interaction Detection}

\author{Junwen Chen\qquad Keiji Yanai\\
Department of Informatics, The University of Electro-Communications, Tokyo, Japan\\
{\tt\small chen-j@mm.inf.uec.ac.jp, yanai@cs.uec.ac.jp}
}
\maketitle

\begin{abstract}
  Human-object interaction (HOI) detection as a downstream of object detection tasks requires localizing pairs of humans and objects and extracting the semantic relationships between humans and objects from an image. 
  Recently, one-stage approaches have become a new trend for this task due to their high efficiency. 
  However, these approaches focus on detecting possible interaction points or filtering human-object pairs, ignoring the variability in the location and size of different objects at spatial scales. 
  To address this problem, we propose a transformer-based method, QAHOI (Query-Based Anchors for Human-Object Interaction detection), which leverages a multi-scale architecture to extract features from different spatial scales and uses query-based anchors to predict all the elements of an HOI instance. 
  We further investigate that a powerful backbone significantly increases accuracy for QAHOI, and QAHOI with a transformer-based backbone outperforms recent state-of-the-art methods by large margins on the HICO-DET benchmark.
  The source code is available at  
  \url{https://github.com/cjw2021/QAHOI}.
\end{abstract}

\section{Introduction}
\label{sec:intro}

Human-object interaction (HOI) detection has recently received increasing attention as a field with great potential applications.
HOI detection approaches need to extract the semantic relationships between humans and objects and predict a set of $\langle$human, object, action$\rangle$ triplets within an image.
Specifically, an HOI instance is a pair of human and object bounding boxes, and a corresponding action class represents the relationship between them.
HOI detection can be seen as a combination of two parts: object detection and human-object interaction recognition.
According to the inference process of these two parts, existing HOI detection approaches can be divided into two-stage and one-stage.

\begin{figure}
  \centering
  \begin{subfigure}{0.49\linewidth}
    \centering
    \includegraphics[height=4.1cm]{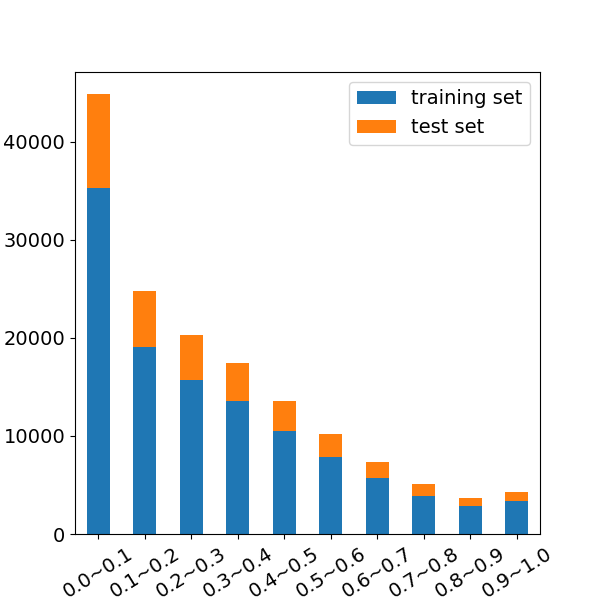}
    \caption{Larger Area}
    \label{fig:f1-a}
  \end{subfigure}
  \begin{subfigure}{0.49\linewidth}
    \centering
    \includegraphics[height=4.1cm]{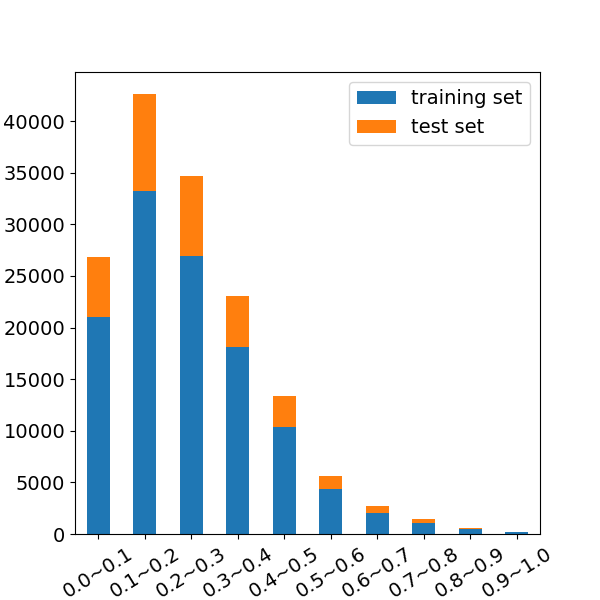}
    \caption{Center Distance}
    \label{fig:f1-b}
  \end{subfigure}
  \caption{The spatial distribution of the HOI instances in the HICO-DET dataset \cite{chao2018learning}. 
  The larger area and center distance indicate the larger area of the human and object bounding box and the distance between the center of the human and object bounding box in an HOI instance, respectively.
  The bounding boxes are normalized according to the image size.}
  \label{fig:HICO-DET-distribution}
\end{figure}
From the beginning to the present, two-stage approaches
\cite{chao2018learning,qi2018learning,gao2018ican,li2019transferable,zhou2019relation,bansal2020detecting,wan2019pose,
   ulutan2020vsgnet,hou2020visual,gao2020drg,zhang2020spatio,zhong2021polysemy,gupta2019no}
as an intuitive approach build the method upon the off-the-shelf object detector \cite{ren2016faster}.
In the first stage, the object detector selects some of high scoring human and object bounding boxes from detection results and extracts appearance features or crops images of the selected bounding boxes.
In the second stage, for each of the human-object pairs, the human and object appearance features are used to predict the action class scores separately or fused with supplementary semantic information as a pairwise interaction feature to predict action class scores directly.
For the two-stage approaches, the difficulty lies in the integration of semantic information of human-object pairs. With the split human-object pairs, spatial information can be extracted to enhance the appearance features
\cite{chao2018learning,gao2018ican,li2019transferable,bansal2020detecting,wan2019pose,ulutan2020vsgnet,
   hou2020visual,gao2020drg,zhang2020spatio,zhong2021polysemy,gupta2019no}.
Furthermore, fine-grained information like human pose can be used as supplementary semantic information \cite{li2019transferable,zhou2019relation,wan2019pose,zhong2021polysemy,gupta2019no}, and graph-based methods \cite{qi2018learning,zhou2019relation,ulutan2020vsgnet,gao2020drg,zhang2020spatio} are also well suited to the understanding of complex semantic relationships between humans and objects.
Without considering the object detection process, the two-stage approach can design a complex model to fuse abundant semantic information of human-object pairs and achieve high accuracy.
However, processing all of the human-object pairs is time-consuming, and the appearance features limited in the bounding box are lack contextual information when the human and the object are far apart.
As shown in Figure~\ref{fig:f1-b}, the spatial distribution of a widely used HOI detection dataset, HICO-DET \cite{chao2018learning}, HOI instances with the center distance between the human and object bounding box more than a third of the image size commonly exist.

To achieve high efficiency, one-stage approaches
\cite{gkioxari2018detecting,liao2020ppdm,kim2021hotr,zou2021end,chen2021reformulating,tamura2021qpic,
   zhong2021glance,kim2020uniondet,wang2020learning}
detect human-object pairs and recognize the corresponding action class in parallel.
A commonly adopted way is to make use of the interaction point, which is between the human-object pair \cite{liao2020ppdm,zhong2021glance,wang2020learning}.
A key-point heatmap prediction network such as Hourglass-104 \cite{newell2016stacked}, or DLA-34 \cite{yu2018deep} is applied to extract appearance features at first.
Then the interaction points, the human and object offset vectors from the interaction points, the action class located at the interaction point, and the human and object boxes are predicted through a multi-branch network.
The interaction points plus the human and object offset vectors indicate where the center points of the human and object are located.
To match the human and object boxes, a matching process is required.
Although interaction points converge the HOI instance detection and recognition together, there are mainly two drawbacks such as the semantic features are ambiguous when the interaction point is far apart from the human and object, and the lack of a multi-scale architecture which is commonly used in object detection.

Both the two-stage and one-stage approaches suffer from the problem of poorly extracted semantic features due to the locality of convolution neural networks (CNNs).
On the other hand, the transformer \cite{waswani2017attention} is well utilized in vision tasks such as image classification \cite{dosovitskiy2021an,liu2021swin}, object detection \cite{liu2021swin,carion2020end,zhu2020deformable}, and semantic segmentation \cite{liu2021swin,wang2020max}.
The self-attention mechanism in the transformer shows encouraging potential for the extraction of contextual visual features, which is beneficial for the HOI detection task.
To extract the semantic features between the human-object pairs with more contextual information and less irrelevant local information, transformer-based HOI detection methods are proposed \cite{kim2021hotr,zou2021end,chen2021reformulating,tamura2021qpic}.
As query embeddings in the transformer decoder represent HOI instances and incorporate object detection and interaction recognition together, the transformer-based HOI detection methods also can be seen as query-based methods which belong to the one-stage approach.
By introducing the transformer as a powerful feature extractor in the HOI detection task, promising results are observed.
However, the transformer-based methods \cite{kim2021hotr,zou2021end,chen2021reformulating,tamura2021qpic} are built upon the CNN backbone, and the multi-head attention used in transformer suffers from a quadratic complexity with the growth of the feature map size.
Thus, these transformer-based methods only use the low-resolution feature map from the CNN backbone and leave a burden for the transformer encoder to extract spatial semantic information.
Besides, the training of the high complexity transformer suffers from slow convergence, and pre-training the model in object detection task and fine-tuning in HOI detection task are always used to obtain a fine result.

\begin{figure*}
  \centering
  \includegraphics[width=0.9\linewidth]{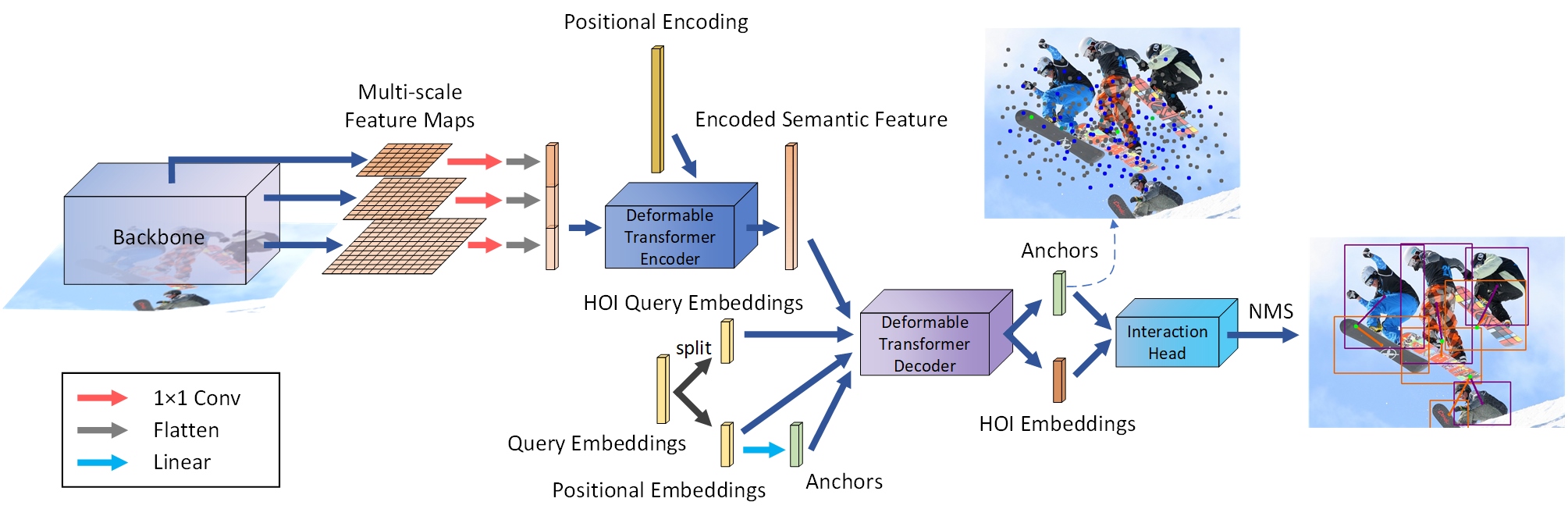}
  \caption{This figure illustrates the overall architecture of the proposed method, QAHOI.
     QAHOI uses a hierarchical backbone and the deformable transformer encoder to extract the semantic feature in a multi-scale manner.
     The deformable transformer decoder is used to decode the HOI embeddings according to the HOI query embeddings and anchors derived from the query embeddings.
     On top of the decoder is an interaction detection head to predict the HOI instance for each anchor with the corresponding HOI embedding.}
  \label{fig:overall}
\end{figure*}
As shown in Figure~\ref{fig:f1-a}, most HOI instances consist of boundary boxes of humans and objects with an area less than 0.1$\times$ image size.
The recent transformer-based one-stage approaches lack a multi-scale architecture to detect the HOI instances containing small objects and are hard to train.
To address these problems above, we proposed a transformer-based method, which leverages a hierarchical backbone to extract multi-scale visual features, and a deformable transformer \cite{zhu2020deformable} to encode the multi-scale semantic features and decode the HOI instances.
The deformable transformer decoder generates the reference points to calculate the multi-scale deformable attention. 
In our case, the reference points in the deformable transformer decoder act as the anchors for aggregating the multi-scale semantic feature from the deformable transformer encoder to the HOI embeddings. With the anchors, an interaction detection head predicts the HOI instances with the HOI embeddings directly.
As the anchors are used throughout the HOI embeddings' decoding and the HOI instances prediction process, we call our method {\bf Q}uery-Based {\bf A}nchors for {\bf HOI} detection, QAHOI.

To summarize, our contributions are three-fold:
\begin{itemize}
  \item We propose a multi-scale transformer-based method, QAHOI for HOI detection, which leverages query-based anchors to extract the HOI embeddings and predict the HOI instances.
  \item We combine a hierarchical backbone with a deformable transformer encoder to build a powerful multi-scale feature extractor, beneficial for the HOI detection task. In addition, we first study and implement the transformer-based backbone on the one-stage HOI detection method and find its great potential for the HOI detection task.
  \item By leveraging the multi-scale architecture, the attention mechanism of the transformer in the whole model, and the flexible query-based anchors, our method outperforms recent state-of-the-art methods by a large margin.
\end{itemize}

\section{Related Work}
\noindent{\bf Two-stage Approaches.}\space The two-stage approach simplifies the HOI detection to an interaction recognition problem with prepared human-object pairs.
The multi-stream architecture was first introduced in the Human-Object Region-based Convolutional Neural Networks (HO-RCNN) \cite{chao2018learning}, which extracts human appearance features, object appearance features, and spatial semantic features separately.
Based on the HO-RCNN, subsequent methods improve the multi-stream architecture by incorporating the advanced feature extracting module \cite{gao2018ican} or the graph module \cite{zhou2019relation, ulutan2020vsgnet, gao2020drg, zhang2020spatio} and fusing the human pose information \cite{li2019transferable, zhong2021polysemy, gupta2019no}.
\\
\noindent{\bf One-stage Approaches.}\space The one-stage approach is proposed with creative designs and adopts a two-branch architecture in general.
The Union-level Detector (UnionDet) \cite{kim2020uniondet} builds a two-branch architecture to detect the union regions of the human-object pairs and the localization of instances in interactions.
The Parallel Point Detection and Matching (PPDM) \cite{liao2020ppdm} defines the interaction point as the midpoint of the human and object center points and matches the human and object instances via interaction points.
The Glance and Gaze Network (GGNet) \cite{zhong2021glance} extends the idea of the interaction point by inferring a set of action-aware points (ActPoints) around each pixel of the feature map.
However, the methods using interaction points \cite{liao2020ppdm,zhong2021glance,wang2020learning} or union regions \cite{kim2020uniondet} require a matching or gathering process to clarify the HOI instance.
On the other hand, the transformer-based methods \cite{kim2021hotr,zou2021end,chen2021reformulating,tamura2021qpic} using the self-attention mechanism of the transformer to extract contextual semantic information and the embeddings to represent the HOI instance become a new trend of the HOI detection task.
Tamura \etal~\cite{tamura2021qpic} transform the object detection head of the transformer-based object detector DETR \cite{carion2020end} into a simple interaction detection head to predict all of the elements of the HOI instance directly.
Similarly, Zou \etal~\cite{zou2021end} combine the CNN backbone and the transformer to predict HOI instances directly from the query embeddings.
Both Chen \etal~\cite{chen2021reformulating} and Kim \etal~\cite{kim2021hotr} propose a transformed-based two-branch architecture, which constructs an instance decoder and an interaction decoder to decode the boxes and action classes of the HOI instances in parallel.

\section{Method}
Our purpose is to address the drawbacks in the recent one-stage approaches that lack a multi-scale architecture and suffers from a poor CNN backbone for the HOI detection task.
The deformable DETR \cite{zhu2020deformable} develops the deformable multi-scale attention module to reduce the complexity of attention in DETR to the linear complexity with the spatial size, which achieves a multi-scale transformer-based object detector.
Our proposed method, QAHOI, further improves this idea to solve HOI detection as a dense prediction problem.
QAHOI adapts the deformable transformer decoder to an HOI instance detector by using the query embeddings to generate anchors and decode the HOI information.
The overall architecture of QAHOI is shown in Figure~\ref{fig:overall}.

\subsection{Multi-Scale Feature Extractor}
For the one-stage approaches based on interaction points or composed with the transformer, a CNN backbone such as Hourglass-104
\cite{newell2016stacked}, DLA-34 \cite{yu2018deep},
ResNet-50 and ResNet-101 \cite{he2016deep} is an ordinary setting. However, these methods ignore two factors of using CNN backbones.
First, the CNN is poor at capturing non-local semantic features like the relationships between humans and objects.
And the way of using the low-resolution feature map with the large receptive field neglects the spatial information on a small scale.
The transformer with the attention mechanism is powerful at extracting semantic information from the image. Thus, recent one-stage methods
\cite{kim2021hotr,zou2021end,chen2021reformulating,tamura2021qpic} normally build a feature extractor consisting of a CNN backbone and a transformer encoder.
To improve the model's expression ability, QAHOI constructs a multi-scale feature extractor by combining a hierarchical backbone and a deformable transformer encoder
\cite{zhu2020deformable} as shown in Figure~\ref{fig:overall}.
The hierarchical backbone extracts four stages' feature maps for the deformable transformer encoder, which is well designed for processing multi-scale feature maps.
Specifically, given an image of size $3\times H \times W$,
QAHOI uses the last three stages' feature maps $x_{1}\in\mathbb{R}^{2C_{s}\times\frac{H}{8}\times\frac{W}{8}}$,
$x_{2}\in\mathbb{R}^{4C_{s}\times\frac{H}{16}\times\frac{W}{16}}$ and $x_{3}\in\mathbb{R}^{8C_{s}\times\frac{H}{32}\times\frac{W}{32}}$ of the backbone.
The $1 \times 1$ convolution is used to project the feature map $x_{1}$, $x_{2}$ and $x_{3}$ from dimension $C_{s}$ to dimension $C_{d}$.
Then, the multi-scale feature maps $x_{1}$, $x_{2}$ and $x_{3}$ are flattened and concatenated to $N_{S}$ vectors with $C_{d}$ dimensions as the input of the deformable transformer encoder, and the fixed positional encoding indicates the scale level of the input.
$N_{S}$ is the sum of pixel numbers of the three feature maps from the backbone.
The deformable transformer encoder extracts the semantic feature $S\in \mathbb{R}^{N_{S}\times C_{d}}$ in a multi-scale manner and provides it for the deformable transform decoder to decode the HOI instances.

\begin{figure}
  \centering
  \begin{tabular}{c}
     \includegraphics[width=0.9\linewidth]{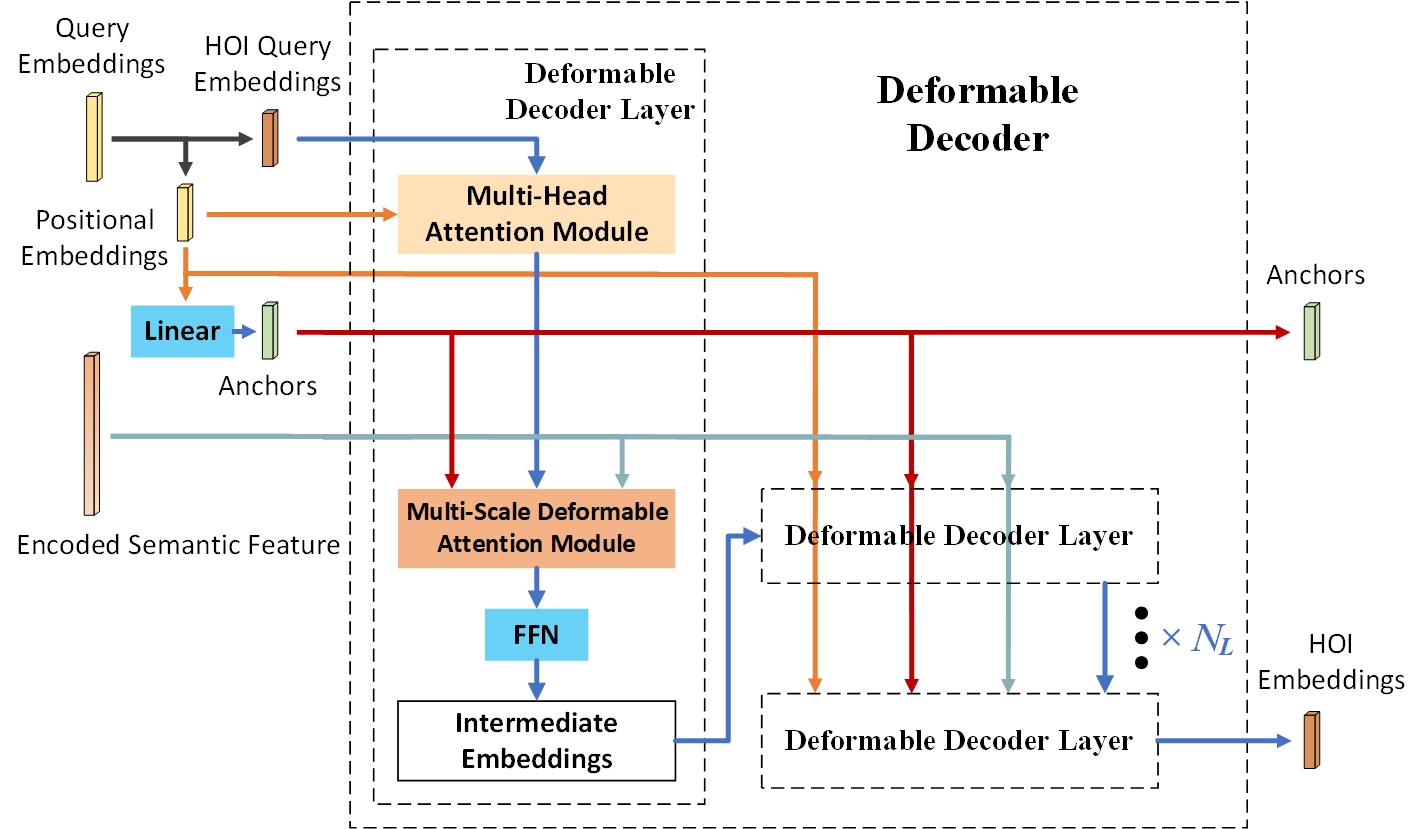}
  \end{tabular}
  \caption{The decoding process of the deformable transformer decoder.}
  \label{fig:decoder}
\end{figure}
\subsection{Predicting HOI with Query-Based Anchors}
According to the deformable DETR,
the query embeddings of the deformable transformer decoder in QAHOI are split equally into two parts,
one as the HOI query embeddings $Q_{HOI}\in \mathbb{R}^{N_{q}\times C_{d}}$ and the other as the positional embeddings $Q_{Pos}\in \mathbb{R}^{N_{q}\times C_{d}}$,
and the anchors $P\in\mathbb{R}^{N_{q}\times 2}$ are generated from positional embeddings $Q_{Pos}$ via a linear layer.
With the HOI query embeddings and the anchors, the HOI embeddings $E\in\mathbb{R}^{N_{q}\times C_{d}}$ are decoded by the deformable transformer decoder's attention mechanism with the source of the encoded semantic feature from the deformable transformer encoder. 
The decoding process of the deformable transformer decoder is shown in Figure \ref{fig:decoder}.
The self-attention of the HOI query embeddings are calculated by the multi-head attention module \cite{waswani2017attention} with the positional embeddings,
and the anchors aggregate the semantic feature from the output of the deformable transform encoder
to calculate the multi-scale deformable attention \cite{zhu2020deformable} with the HOI query embeddings.
Besides, after the calculation of the multi-scale deformable attention,
a feed-forward network (FFN) composed of linear layers is used to process the output embeddings.
The self-attention and the multi-scale attention are calculated in the stacked decode layer for $N_{L}$ times,
and the last layer outputs the HOI embeddings for the interaction detection head to predict the HOI instances.

\begin{figure}
  \centering
  \begin{tabular}{c}
     \includegraphics[width=0.8\linewidth]{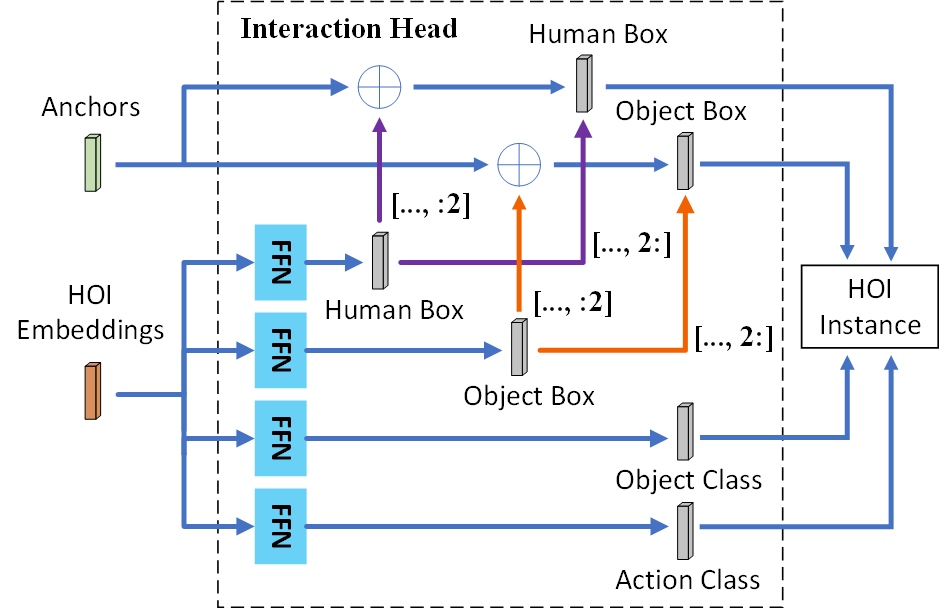}
  \end{tabular}
  \caption{The interaction head predicts the HOI instances based on the anchors.}
  \label{fig:head}
\end{figure}

QAHOI implements a simple interaction head which is similar to the QPIC \cite{tamura2021qpic},
and the difference is that QAHOI combines each HOI embedding with a certain anchor.
Hence, QAHOI feeds the decoded HOI embeddings into the interaction head to predict the HOI instances based on the anchors.
Figure~\ref{fig:head} shows the predicting process of the interaction head in QAHOI.
Following the deformable DETR, each anchor $(p_{x}, p_{y})$ of the anchor set $P\in\mathbb{R}^{N_{q}\times 2}$
acts as the base point for the bounding boxes of a pair of a human and an object.
Thus, the human and object boxes $B^{h}$, $B^{o}\in\mathbb{R}^{N_{q}\times 4}$ predicted by the FFN in the interaction head
are composed of $\{d_{x}$, $d_{y}$, $w$, $h\}$,
where $d_{x}$ and $d_{y}$ denote the offsets between the anchor and the box's center, and $w$ and $h$ denote the box's width and height.
Then, the final bounding boxes $\hat{B}^{h}$, $\hat{B}^{o}$ are composed of $\{d_{x}+p_{x}$, $d_{y}+p_{y}$, $w$, $h\}$.
Finally, the object class of the object boxes $O\in\mathbb{R}^{N_{q}\times K_{o}}$
and the action class of the HOI instances $A\in\mathbb{R}^{N_{q}\times K_{a}}$
are combined with the human and object bounding boxes $\hat{B}^{h}$, $\hat{B}^{o}$ to construct the output HOI instances.

\begin{table*}
  \centering
  \resizebox{\linewidth}{!}{%
     \begin{tabular}{cccccccccccc}
        \hline
        \multicolumn{1}{c|}{}                                                         & \multicolumn{1}{c|}{}                                     & \multicolumn{1}{c|}{}                                         & \multicolumn{1}{c|}{}                            & \multicolumn{1}{c|}{Fine-tuned} & \multicolumn{1}{c|}{}        & \multicolumn{3}{c|}{Default} & \multicolumn{3}{c}{Known Object} \\
        \multicolumn{1}{c|}{}                                                         & \multicolumn{1}{c|}{Architecture}                         & \multicolumn{1}{c|}{Method}                                   & \multicolumn{1}{c|}{Backbone}                    & \multicolumn{1}{c|}{Detection}  & \multicolumn{1}{c|}{Feature} & \textit{Full}                & \textit{Rare}    & \multicolumn{1}{c|}{\textit{Non-Rare}} & \textit{Full}    & \textit{Rare}    & \textit{Non-Rare} \\ \hline \hline
        \multicolumn{1}{c|}{\multirow{10}{*}{\rotatebox[origin=c]{90}{Two-Stage}}}    & \multicolumn{1}{c|}{\multirow{6}{*}{Multi-Stream}}        & \multicolumn{1}{l|}{No-Frills \cite{gupta2019no}}             & \multicolumn{1}{c|}{ResNet-152}                  & \multicolumn{1}{c|}{\ding{55}}  & \multicolumn{1}{c|}{A+S+P}   & 17.18                        & 12.17            & \multicolumn{1}{c|}{18.68}             & -                & -                & -                 \\
        \multicolumn{1}{c|}{}                                                         & \multicolumn{1}{c|}{}                                     & \multicolumn{1}{l|}{PMFNet \cite{wan2019pose}}                & \multicolumn{1}{c|}{ResNet-50-FPN}               & \multicolumn{1}{c|}{\ding{55}}  & \multicolumn{1}{c|}{A+S}     & 17.46                        & 15.65            & \multicolumn{1}{c|}{18.00}             & 20.34            & 17.47            & 21.20             \\
        \multicolumn{1}{c|}{}                                                         & \multicolumn{1}{c|}{}                                     & \multicolumn{1}{l|}{Bansal \etal~\cite{bansal2020detecting}} & \multicolumn{1}{c|}{ResNet-101}                  & \multicolumn{1}{c|}{\ding{51}}  & \multicolumn{1}{c|}{A+S+L}   & 21.96                        & 16.43            & \multicolumn{1}{c|}{23.62}             & -                & -                & -                 \\
        \multicolumn{1}{c|}{}                                                         & \multicolumn{1}{c|}{}                                     & \multicolumn{1}{l|}{PD-Net \cite{zhong2021polysemy}}          & \multicolumn{1}{c|}{ResNet-152}                  & \multicolumn{1}{c|}{\ding{55}}  & \multicolumn{1}{c|}{A+S+P+L} & 22.37                        & 17.61            & \multicolumn{1}{c|}{23.79}             & 26.86            & 21.70            & 28.44             \\
        \multicolumn{1}{c|}{}                                                         & \multicolumn{1}{c|}{}                                     & \multicolumn{1}{l|}{VCL \cite{hou2020visual}}                 & \multicolumn{1}{c|}{ResNet-50}                   & \multicolumn{1}{c|}{\ding{51}}  & \multicolumn{1}{c|}{A+S}     & 23.63                        & 17.21            & \multicolumn{1}{c|}{25.55}             & 25.98            & 19.12            & 28.03             \\ \cline{2-12}
        \multicolumn{1}{c|}{}                                                         & \multicolumn{1}{c|}{\multirow{4}{*}{Graph-Based}}         & \multicolumn{1}{l|}{RPNN \cite{zhou2019relation}}             & \multicolumn{1}{c|}{ResNet-50}                   & \multicolumn{1}{c|}{\ding{55}}  & \multicolumn{1}{c|}{A+P}     & 17.35                        & 12.78            & \multicolumn{1}{c|}{18.71}             & -                & -                & -                 \\
        \multicolumn{1}{c|}{}                                                         & \multicolumn{1}{c|}{}                                     & \multicolumn{1}{l|}{VSGNet \cite{ulutan2020vsgnet}}           & \multicolumn{1}{c|}{ResNet-152}                  & \multicolumn{1}{c|}{\ding{55}}  & \multicolumn{1}{c|}{A+S}     & 19.80                        & 16.05            & \multicolumn{1}{c|}{20.91}             & -                & -                & -                 \\
        \multicolumn{1}{c|}{}                                                         & \multicolumn{1}{c|}{}                                     & \multicolumn{1}{l|}{DRG \cite{gao2020drg}}                    & \multicolumn{1}{c|}{ResNet-50-FPN}               & \multicolumn{1}{c|}{\ding{51}}  & \multicolumn{1}{c|}{A+S+L}   & 24.53                        & 19.47            & \multicolumn{1}{c|}{26.04}             & 27.98            & 23.11            & 29.43             \\
        \multicolumn{1}{c|}{}                                                         & \multicolumn{1}{c|}{}                                     & \multicolumn{1}{l|}{Zhang \etal~\cite{zhang2020spatio}}      & \multicolumn{1}{c|}{ResNet-50-FPN}               & \multicolumn{1}{c|}{\ding{51}}  & \multicolumn{1}{c|}{A+S}     & 31.33                        & 24.72            & \multicolumn{1}{c|}{33.31}             & 34.37            & 27.18            & 36.52             \\ \hline \hline
        \multicolumn{1}{c|}{\multirow{13}{*}{\rotatebox[origin=c]{90}{one-stage}}}    & \multicolumn{1}{c|}{\multirow{3}{*}{Interaction Points}}  & \multicolumn{1}{l|}{IP-Net \cite{wang2020learning}}           & \multicolumn{1}{c|}{ResNet-50-FPN}               & \multicolumn{1}{c|}{\ding{55}}  & \multicolumn{1}{c|}{A}       & 19.56                        & 12.79            & \multicolumn{1}{c|}{21.58}             & 22.05            & 15.77            & 23.92             \\
        \multicolumn{1}{c|}{}                                                         & \multicolumn{1}{c|}{}                                     & \multicolumn{1}{l|}{PPDM \cite{liao2020ppdm}}                 & \multicolumn{1}{c|}{Hourglass-104}               & \multicolumn{1}{c|}{\ding{51}}  & \multicolumn{1}{c|}{A}       & 21.73                        & 13.78            & \multicolumn{1}{c|}{24.10}             & 24.58            & 16.65            & 26.84             \\
        \multicolumn{1}{c|}{}                                                         & \multicolumn{1}{c|}{}                                     & \multicolumn{1}{l|}{GGNet \cite{zhong2021glance}}             & \multicolumn{1}{c|}{Hourglass-104}               & \multicolumn{1}{c|}{\ding{51}}  & \multicolumn{1}{c|}{A}       & 23.47                        & 16.48            & \multicolumn{1}{c|}{25.60}             & 27.36            & 20.23            & 29.48             \\ \cline{2-12}
        \multicolumn{1}{c|}{}                                                         & \multicolumn{1}{c|}{\multirow{10}{*}{Transformer-Based}}  & \multicolumn{1}{l|}{HOITrans \cite{zou2021end}}               & \multicolumn{1}{c|}{ResNet-101}                  & \multicolumn{1}{c|}{\ding{51}}  & \multicolumn{1}{c|}{A}       & 26.61                        & 19.15            & \multicolumn{1}{c|}{28.84}             & 29.13            & 20.98            & 31.57             \\ 
        \multicolumn{1}{c|}{}                                                         & \multicolumn{1}{c|}{}                                     & \multicolumn{1}{l|}{HOTR \cite{kim2021hotr}}                  & \multicolumn{1}{c|}{ResNet-50}                   & \multicolumn{1}{c|}{\ding{55}}  & \multicolumn{1}{c|}{A}       & 23.46                        & 16.21            & \multicolumn{1}{c|}{25.65}             & -                & -                & -                 \\
        \multicolumn{1}{c|}{}                                                         & \multicolumn{1}{c|}{}                                     & \multicolumn{1}{l|}{HOTR \cite{kim2021hotr}}                  & \multicolumn{1}{c|}{ResNet-50}                   & \multicolumn{1}{c|}{\ding{51}}  & \multicolumn{1}{c|}{A}       & 25.10                        & 17.34            & \multicolumn{1}{c|}{27.42}             & -                & -                & -                 \\ 
        \multicolumn{1}{c|}{}                                                         & \multicolumn{1}{c|}{}                                     & \multicolumn{1}{l|}{AS-Net \cite{chen2021reformulating}}      & \multicolumn{1}{c|}{ResNet-50}                   & \multicolumn{1}{c|}{\ding{55}}  & \multicolumn{1}{c|}{A}       & 24.40                        & 22.39            & \multicolumn{1}{c|}{25.01}             & 27.41            & 25.44            & 28.00             \\
        \multicolumn{1}{c|}{}                                                         & \multicolumn{1}{c|}{}                                     & \multicolumn{1}{l|}{AS-Net \cite{chen2021reformulating}}      & \multicolumn{1}{c|}{ResNet-50}                   & \multicolumn{1}{c|}{\ding{51}}  & \multicolumn{1}{c|}{A}       & 28.87                        & 24.25            & \multicolumn{1}{c|}{30.25}             & 31.74            & 27.07            & 33.14             \\ 
        \multicolumn{1}{c|}{}                                                         & \multicolumn{1}{c|}{}                                     & \multicolumn{1}{l|}{QPIC \cite{tamura2021qpic}}               & \multicolumn{1}{c|}{ResNet-101}                  & \multicolumn{1}{c|}{\ding{51}}  & \multicolumn{1}{c|}{A}       & 29.90                        & 23.92            & \multicolumn{1}{c|}{31.69}             & 32.38            & 26.06            & 34.27             \\ \cline{3-12}
        \multicolumn{1}{c|}{}                                                         & \multicolumn{1}{c|}{}                                     & \multicolumn{1}{l|}{\textbf{QAHOI}}                           & \multicolumn{1}{c|}{\textbf{Swin-Tiny}}          & \multicolumn{1}{c|}{\ding{55}}  & \multicolumn{1}{c|}{A}       & 28.47                        & 22.44            & \multicolumn{1}{c|}{30.27}             & 30.99            & 24.83            & 32.84             \\
        \multicolumn{1}{c|}{}                                                         & \multicolumn{1}{c|}{}                                     & \multicolumn{1}{l|}{\textbf{QAHOI}}                           & \multicolumn{1}{c|}{\textbf{Swin-Base}}          & \multicolumn{1}{c|}{\ding{55}}  & \multicolumn{1}{c|}{A}       & 29.47                        & 22.24            & \multicolumn{1}{c|}{31.63}             & 31.45            & 24.00            & 33.68             \\
        \multicolumn{1}{c|}{}                                                         & \multicolumn{1}{c|}{}                                     & \multicolumn{1}{l|}{\textbf{QAHOI}}                           & \multicolumn{1}{c|}{\textbf{Swin-Base}$^{*+}$}   & \multicolumn{1}{c|}{\ding{55}}  & \multicolumn{1}{c|}{A}       & 33.58                        & 25.86            & \multicolumn{1}{c|}{35.88}             & 35.34            & 27.24            & 37.76             \\
        \multicolumn{1}{c|}{}                                                         & \multicolumn{1}{c|}{}                                     & \multicolumn{1}{l|}{\textbf{QAHOI}}                           & \multicolumn{1}{c|}{\textbf{Swin-Large}$^{*+}$}  & \multicolumn{1}{c|}{\ding{55}}  & \multicolumn{1}{c|}{A}       & {\textbf{35.78}}             & {\textbf{29.80}} & \multicolumn{1}{c|}{\textbf{37.56}}    & {\textbf{37.59}} & {\textbf{31.66}} & {\textbf{39.36}}  \\ \hline
     \end{tabular}%
  }
  \caption{Comparison with state-of-the-art on HICO-DET.
     The features of 'A', 'S', 'P' and 'L' represent the appearance feature, spatial feature,
     human pose feature, and language feature, respectively.
     For two-stage approaches, using fine-tuned detection means using a detector \cite{ren2016faster} first trained on the MS-COCO dataset,
     and then fine-tuned on the HICO-DET dataset.
     For one-stage approaches, using fine-tuned detection means initializing the weights of the detection part from a model pre-trained on the MS-COCO dataset and fine-tuning the whole network on the HICO-DET dataset.
     The Swin-Base and Swin-Large backbone with the $^*$ and $^+$ are pre-trained on ImageNet-22K with 384 $\times$ 384 input resolution.
  }
  \label{tab:result-table}
\end{table*}
\subsection{Training and Inference}
Because the human and object are predicted in pairs, the matching process, which is important in the method based on interaction points, is not needed.
Following the training procedure of the QPIC \cite{tamura2021qpic}, the ground-truth set is padded with $\phi$ (no pairs) to the size of $N_{q}$,
and the Hungarian algorithm \cite{kuhn1955hungarian} is used to match all of the $N_{q}$ predictions with the ground-truth set.
For the loss calculated on the matched pairs, the QPIC's loss function is based on the DETR \cite{carion2020end},
and because QAHOI implements the deformable DETR \cite{zhu2020deformable},
we follow the Deformable DETR to calculate the Focal Loss \cite{lin2017focal} of the object class, which is different from the QPIC.
For the anchors derived from the query embeddings, because the query embeddings are learnable parameters, the positions of the anchors are learned during training and fixed during inference.

\subsection{Top K Scores and HOI NMS}
QAHOI requires sufficient anchors to extract multi-scale features. 
In general, the number of anchors far exceeds the number of HOI instances in an image. For the HICO-DET dataset, $96\%$ of the images contains less than 10 HOI instances. 
QAHOI filters the results in two steps. Firstly, the HOI instances with the top $N_{t}$ object class scores are selected. 
Then, an HOI Non-Maximal Suppression (NMS) is used to filter out the final results. The HOI NMS is calculated based on the IoU of humans and objects between HOI instances and the HOI score.
The HOI score is obtained by multiplying the object score and the action score, $c_{\text{HOI}}=c_{o}\cdot c_{a}$.
And a combined IoU of human and object between an HOI instance $i$ and $j$ is calculated as:
\begin{equation}
  \text{IoU}(i, j) = \text{IoU}(B^{(h)}_{i},B^{(h)}_{j})\cdot \text{IoU}(B^{(o)}_{j},B^{(o)}_{j})
  \label{eq:nms}
\end{equation}
The same as the object detection task, a threshold $\delta$ is used to remove HOI instances with low scores for each action category based on the IoU.

\begin{table*}
  \centering
  \begin{minipage}[t]{0.63\linewidth}
    \vspace{0pt}
    \centering
    \resizebox{0.9\linewidth}{!}{
      \begin{tabular}{@{}cccccccc@{}}
        \hline
        \multicolumn{1}{c|}{}                       & \multicolumn{1}{c|}{}       & \multicolumn{1}{c|}{}          & \multicolumn{1}{c|}{Fine-tuned} & \multicolumn{1}{c|}{}                  & \multicolumn{3}{c}{Default} \\
        \multicolumn{1}{c|}{Arch.}                  & \multicolumn{1}{c|}{Method} & \multicolumn{1}{c|}{Backbone}  & \multicolumn{1}{c|}{Detection}  & \multicolumn{1}{c|}{Multi-scale}       & \textit{Full} & \textit{Rare} & \textit{Non-Rare} \\ \hline \hline
        \multicolumn{1}{c|}{\multirow{3}{*}{QPIC}}  & \multicolumn{1}{c|}{(1)}    & \multicolumn{1}{c|}{ResNet-50} & \multicolumn{1}{c|}{\ding{55}}  & \multicolumn{1}{c|}{$x_3$}             & 24.21 & 17.51 & 26.21   \\
        \multicolumn{1}{c|}{}                       & \multicolumn{1}{c|}{(2)}    & \multicolumn{1}{c|}{ResNet-50} & \multicolumn{1}{c|}{\ding{51}}  & \multicolumn{1}{c|}{$x_3$}             & 29.07 & 21.85 & 31.23   \\ \cline{3-8}
        \multicolumn{1}{c|}{}                       & \multicolumn{1}{c|}{(3)}    & \multicolumn{1}{c|}{Swin-Tiny} & \multicolumn{1}{c|}{\ding{55}}  & \multicolumn{1}{c|}{$x_3$}             & 27.19 & 21.32 & 28.95   \\ \hline
        \multicolumn{1}{c|}{\multirow{6}{*}{QAHOI}} & \multicolumn{1}{c|}{(4)}    & \multicolumn{1}{c|}{ResNet-50} & \multicolumn{1}{c|}{\ding{55}}  & \multicolumn{1}{c|}{$x_1,x_2,x_3,x_4$} & 24.35 & 16.18 & 26.80   \\
        \multicolumn{1}{c|}{}                       & \multicolumn{1}{c|}{(5)}    & \multicolumn{1}{c|}{ResNet-50} & \multicolumn{1}{c|}{\ding{51}}  & \multicolumn{1}{c|}{$x_1,x_2,x_3,x_4$} & 26.18 & 18.06 & 28.61   \\ \cline{3-8}
        \multicolumn{1}{c|}{}                       & \multicolumn{1}{c|}{(6)}    & \multicolumn{1}{c|}{Swin-Tiny} & \multicolumn{1}{c|}{\ding{55}}  & \multicolumn{1}{c|}{$x_1,x_2,x_3,x_4$} & 28.09 & 21.65 & 30.01   \\
        \multicolumn{1}{c|}{}                       & \multicolumn{1}{c|}{(7)}    & \multicolumn{1}{c|}{Swin-Tiny} & \multicolumn{1}{c|}{\ding{55}}  & \multicolumn{1}{c|}{$x_1,x_2,x_3$}     & 28.47 & 22.44 & 30.27   \\
        \multicolumn{1}{c|}{}                       & \multicolumn{1}{c|}{(8)}    & \multicolumn{1}{c|}{Swin-Tiny} & \multicolumn{1}{c|}{\ding{55}}  & \multicolumn{1}{c|}{$x_2,x_3$}         & 28.12 & 20.43 & 30.41   \\
        \multicolumn{1}{c|}{}                       & \multicolumn{1}{c|}{(9)}    & \multicolumn{1}{c|}{Swin-Tiny} & \multicolumn{1}{c|}{\ding{55}}  & \multicolumn{1}{c|}{$x_3$}             & 26.65 & 19.13 & 28.89   \\ \hline
      \end{tabular}
    }
    \caption{Evaluations of the training strategies and the effect of multi-scale feature maps and transformer-based backbone.}
    \label{tab:ablation-training}
    \vspace{10pt}
    \resizebox{0.6\linewidth}{!}{
      \begin{tabular}{@{}cccc@{}}
        \hline
        \multicolumn{1}{c|}{}                            & \multicolumn{3}{c}{Default} \\
        \multicolumn{1}{c|}{method}                      & \textit{Full} & \textit{Rare} & \textit{Non-Rare} \\ \hline \hline
        \multicolumn{1}{c|}{base}                        & 26.64         & 20.62         & 28.44             \\
        \multicolumn{1}{c|}{+ topk scores ($N_{t}=100$)} & 26.70         & 20.89         & 28.43             \\
        \multicolumn{1}{c|}{+ NMS ($\delta=0.5$)}        & 28.47         & 22.44         & 30.27             \\ \hline
      \end{tabular}
    }
    \caption{Ablation study of the filtering steps. QAHOI with Swin-Tiny is used as the base method.}
    \label{tab:ablation-filter}
  \end{minipage}\hfill%
  \begin{minipage}[t]{0.33\linewidth}
    \vspace{0pt}
    \centering
    \begin{subtable}[t]{\linewidth}
      \centering
      \resizebox{0.85\linewidth}{!}{
        \begin{tabular}{@{}cccc@{}}
          \hline
          \multicolumn{1}{c|}{}                   & \multicolumn{3}{c}{$N_{t}$} \\
          \multicolumn{1}{c|}{topk scores}        & 50    & 100         & 150   \\ \hline \hline
          \multicolumn{1}{c|}{$c_{a}$}            & 26.63 & 26.63       & 26.63 \\ 
          \multicolumn{1}{c|}{$c_{o}$}            & 26.69 & {\bf 26.70} & 26.64 \\
          \multicolumn{1}{c|}{$c_{a}\cdot c_{o}$} & 26.63 & 26.63       & 26.63 \\ \hline
        \end{tabular}
      }
      \caption{Ablation study of the top K scores. Three kinds of top k scores with three top k numbers are tested without the NMS process.}
      \label{tab:ablation-topk_scores}
    \end{subtable}
    \begin{subtable}[t]{\linewidth}
      \vspace{10pt}
      \centering
      \resizebox{0.9\linewidth}{!}{
        \begin{tabular}{@{}ccccc@{}}
          \hline
          \multicolumn{1}{c|}{}                                  & \multicolumn{4}{c}{IoU threshold}        \\
          \multicolumn{1}{c|}{$\text{IoU}$}                      & 0.4   & 0.5         & 0.6    & 0.7    \\ \hline \hline
          \multicolumn{1}{c|}{$\text{IoU}^{h}$}                  & 27.85 & 27.93       & 27.96  & 27.93  \\
          \multicolumn{1}{c|}{$\text{IoU}^{o}$}                  & 26.69 & 26.77       & 26.84  & 26.85  \\
          \multicolumn{1}{c|}{$\text{IoU}^{h}\cdot\text{IoU}^o$} & 28.41 & {\bf 28.47} & 28.37  & 28.07  \\ \hline
        \end{tabular}
      }
      \caption{Ablation study of the NMS method. Three kinds of IoU calculation methods with four thresholds are tested, the top k score $c_{o}$ is used, and $N_{t}=100$.}
      \label{tab:ablation-nms}
    \end{subtable}
    \caption{Ablation studies of top K scores and NMS methods. QAHOI with Swin-Tiny is used to conduct results on the {\em Full} category of the HICO-DET dataset.}
    \label{tab:ablation-nms_topk}
  \end{minipage}
\end{table*}
\section{Experiments}
\subsection{Experimental Setting}
\noindent{\bf Dataset.}\space We conduct the experiments on the HICO-DET \cite{chao2018learning} dataset,
which contains 47,776 images (38,118 in the training set and 9,658 in the test set).
HICO-DET has 117 action classes and 80 object classes (the object classes same as the MS-COCO \cite{lin2014microsoft} dataset),
and the action classes and the object classes constitute 600 HOI classes.
Based on the number of instances of the 600 HOI classes in the dataset, these HOI classes are divided into three categories:
{\em Full} (all of the HOI classes), {\em Rare} (138 classes with less than 10 instances), and {\em Non-Rare} (462 classes with 10 or more than 10 instances).
We report the results (in Table~\ref{tab:result-table}) on the Default setting (with unknown objects) and the Known Object setting (without unknown objects) of the HICO-DET.
\\
{\bf Metric.}\space The mean average precious (mAP) is used to evaluate the predicted HOI instances.
For a true positive HOI instance,
the intersection over union (IoU) between the predicted human bounding box and the ground-truth human bounding box is higher than 0.5,
and the IoU between the predicted object and the ground-truth object bounding box is also higher than 0.5.
As usual, we report the mAP on the {\em Full}, {\em Rare}, and {\em Non-Rare} categories of the HICO-DET.
\\
{\bf Implementation Details.}\space For the backbone, we train QAHOI with Swin-Transformer \cite{liu2021swin} pre-trained on ImageNet \cite{deng2009imagenet} as our best model.
Specifically, we use Swin-Tiny and Swin-Base pre-trained on ImageNet-1K, and Swin-Base and Swin-Large pre-trained on ImageNet-22K.
Following the setting of the Deformable DETR, the deformable transformer encoder and decoder both have 6 layers ($N_{L}=6$),
the number of the query embeddings is $N_{q}=300$, and top $N_{t}=100$ HOI instances are selected by object scores.
In the NMS process, $\delta=0.5$ is used to filter the HOI instances by the combined IoU.
For the Swin-Tiny, Swin-Base and Swin-Large as the backbone, the first stage's feature map's dimensions are $C_{s}=96$, $C_{s}=128$ and $C_{s}=192$.
Following the deformable DETR's setting, the dimension of the embeddings in the deformable transformer is $C_{d}=256$.
We use the AdamW \cite{loshchilov2018decoupled} optimizer with the backbone's learning rate of $10^{-5}$ and other's $10^{-4}$, and the weight decay of $10^{-4}$. 
We train the model for 150 epochs with a batch size of 16 (two images per GPU, 8 GPUs),
and the learning rates of the backbone and others are decayed at 120 epochs.

\subsection{Comparison with State-of-the-Arts}
The results compared with the state-of-the-art methods on the HICO-DET are shown in Table~\ref{tab:result-table}.
We use QAHOI with the Swin Transformer as our best model to compare with other state-of-the-art methods.
Compared with the two-stage approaches,
by leveraging the attention mechanism to extract semantic information, QAHOI with Swin-Large exceeds the state-of-the-art graph-based method \cite{zhang2020spatio} by 4.45 mAP (relatively 14.2$\%$).
Compared with the one-stage approaches,
with the multi-scale feature maps and multi-scale deformable attention, even we do not train a detector on the MS-COCO dataset, which is beneficial for the object detection part of the model,
QAHOI with Swin-Large backbone still outperforms the state-of-the-art one-stage method, QPIC with 5.88 mAP (relatively 19.7$\%$).
We found that the better the performance of the pre-trained backbone in the classification task became, the further improvement in accuracy we achieved in the HOI detection.
The mAP of QAHOI with Swin-Base backbone pre-trained on ImageNet-20K is 4.1 (relatively 13.9$\%$) higher than the same backbone pre-trained on ImageNet-1K.

\begin{figure*}
  \centering
  \begin{subfigure}{0.48\linewidth}
    \centering
    \includegraphics[height=3.7cm]{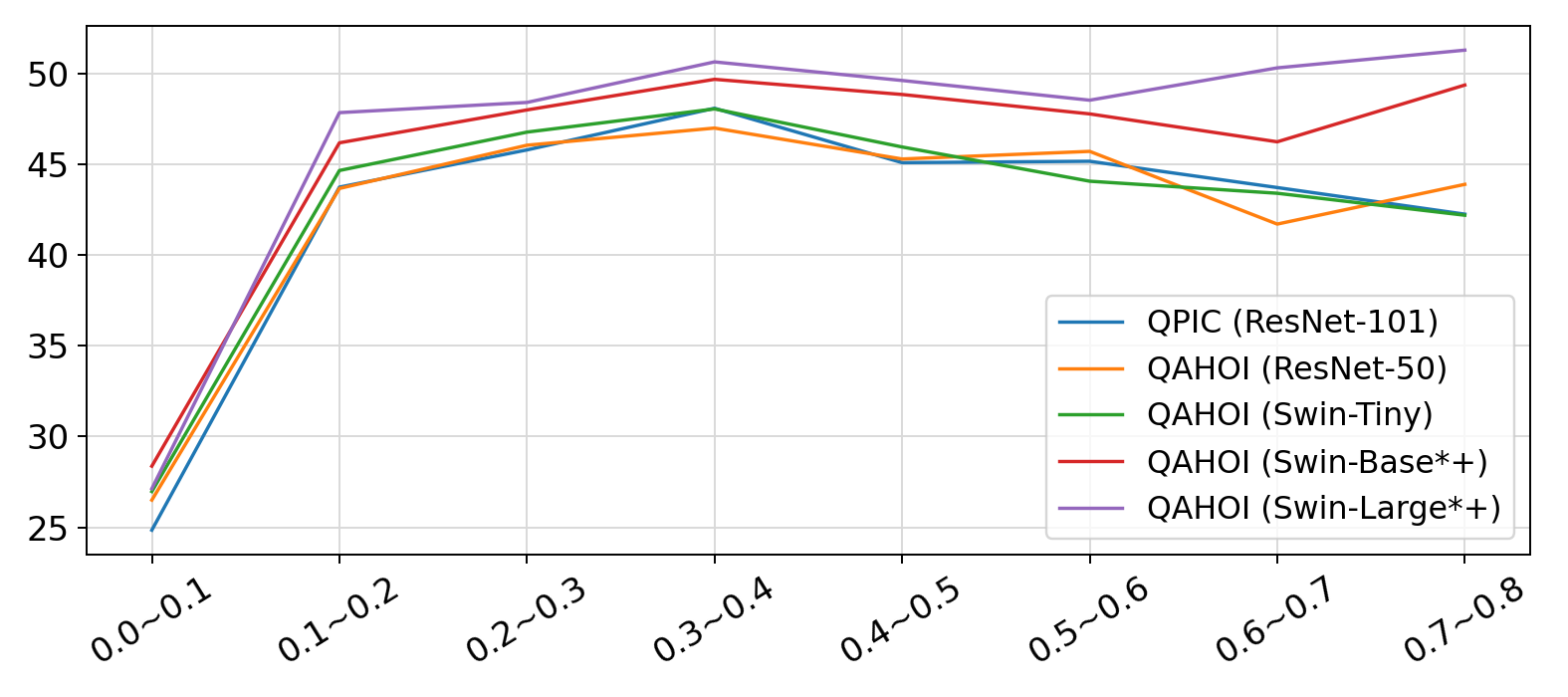}
    \caption{AP results on different large areas.}
    \label{fig:spatial_area}
  \end{subfigure}
  \begin{subfigure}{0.48\linewidth}
    \centering
    \includegraphics[height=3.7cm]{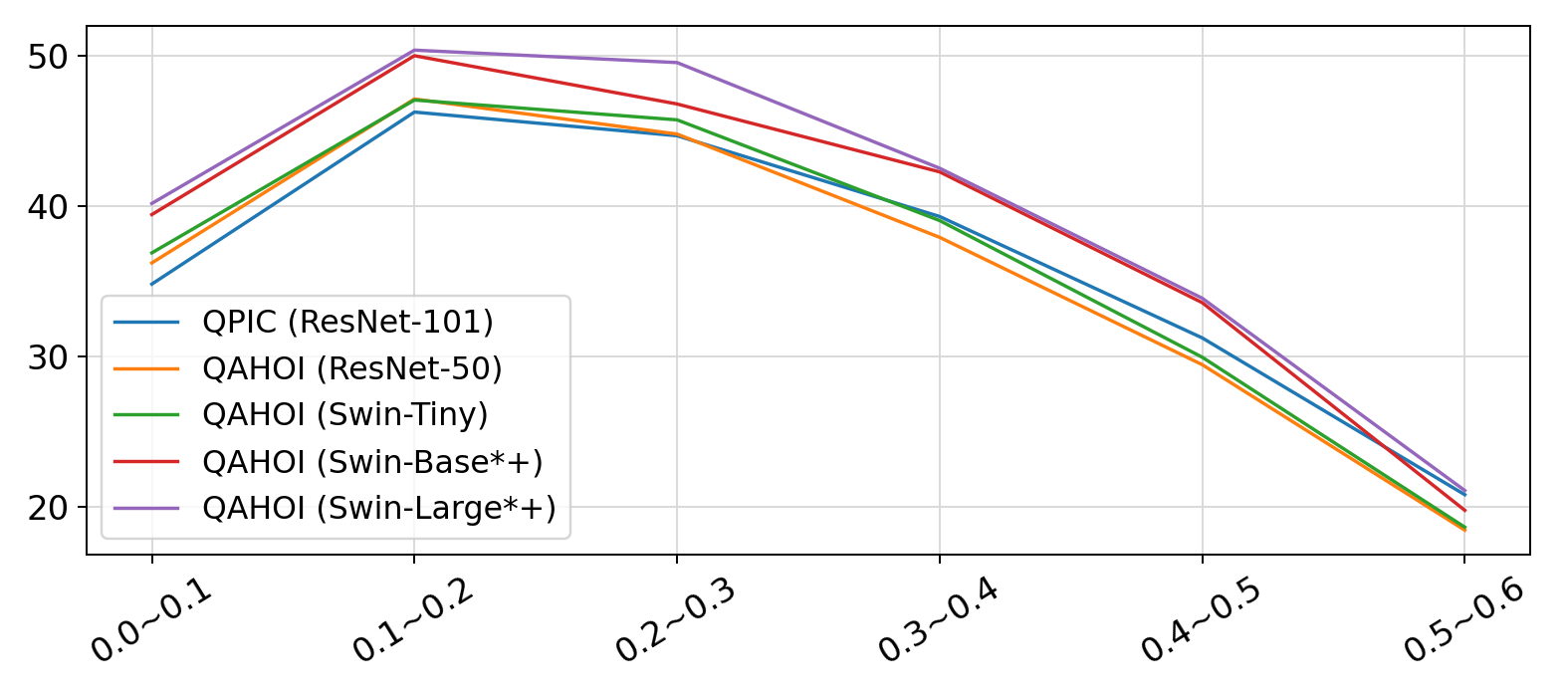}
    \caption{AP results on different center distances.}
    \label{fig:spatial_distance}
  \end{subfigure}
  \caption{Evaluations on different spatial scales of HOI instances.}
  \label{fig:results_in_spatial}
\end{figure*}
\subsection{Ablation Study}
We conduct ablation studies using CNN-based and Transformer-based backbones. 
For the CNN-based backbone, we use ResNet-50 and investigate the performance of two training strategies, starting from scratch and fine-tuning the weights of the detector. 

\noindent{\bf Training strategies.}\space The same as QPIC, we use the deformable DETR's weight which is trained on the MS-COCO dataset, to initialize QAHOI and then fine-tune QAHOI on the HICO-DET dataset.
Following the deformable DETR's implementation,
An additional low-resolution feature map $x_{4}\in\mathbb{R}^{C_{d}\times\frac{H}{64}\times\frac{W}{64}}$ is generated
by using a $3 \times 3$ convolution on the feature map $x_3$.
The additional feature map $x_{4}$ has the dimension $C_{d}=256$, which is the same as the embeddings in the deformable transformer.
We also train QAHOI and QPIC with ResNet-50 and Swin-Tiny from scratch, respectively.
From the results in Table~\ref{tab:ablation-training}, without training a detector, (4) QAHOI with ResNet-50 or (7) Swin-Tiny achieves better results on the {\em Full} and {\em Non-Rare} categories compared with (1) QPIC with ResNet-50 or (3) Swin-Tiny.
\\
\noindent{\bf Multi-scale feature maps.}\space We use the Swin-Tiny backbone to investigate the effect of different combinations of feature maps on the accuracy of the proposed method.
From the results in Table~\ref{tab:ablation-training}(6), the additional feature map does not improve the accuracy. 
For methods (7)(8)(9) of QAHOI, the accuracy decreases with the removal of multi-scale feature maps.
Comparing (9) to (7), using the feature maps of three stages gives a model accuracy improvement of 1.82 mAP (relatively 6.8$\%$) on the {\em Full} category.
\\
\noindent{\bf CNN-based backbone vs Transformer-based backbone.}\space The Swin-Tiny has the model size and the computation complexity similar to ResNet-50, but the accuracy on ImageNet is higher than ResNet-50.
Without training an object detector, compared with the model trained with ResNet-50 in Table~\ref{tab:ablation-training}(1)(4), the transformer-based backbone Swin-Tiny improves the accuracy of both (3) QPIC (2.98 mAP, relatively 12.3$\%$) and (7) QAHOI (4.12 mAP, relatively 16.9$\%$), and (7) QAHOI with Swin-Tiny is better than (3) QPIC with Swin-Tiny both of the accuracy and improvement, which means our method has a great potential based on well-designed backbones.
The results of QAHOI trained with Swin-Base and Swin-Large in Table~\ref{tab:result-table} also show that using a backbone with higher accuracy on classification tasks can improve the accuracy of HOI detection significantly.
The result of (5) QAHOI fine-tuned from Deformable DETR is lower than (2) QPIC fine-tuned from DETR. 
One of the reasons is that QPIC uses the DETR with 500 epochs of training, while we use the deformable DETR with only 50 epochs of training. 
QAHOI would have achieved better results if we have fine-tuned the deformable DETR with more epochs.
\\
\noindent{\bf Top K scores and HOI NMS.}\space The filtering process is important to QAHOI, in Table~\ref{tab:ablation-filter}, the top K scores step and NMS step improve the accuracy on the {\em Full} category by 1.83 mAP.
To optimize the top K scores step, we test different kinds of scores and top K numbers. 
The results in Table~\ref{tab:ablation-topk_scores} show that using the object score is better than using the action score, and the results of $c_{a}$ and $c_{a}\cdot c_{o}$ are the same, which means the action score is not sensitive to the number of top K.
The best result is obtained under the condition that the top 100 HOI instances of the object score are used as output.
And we further test the IoU calculation and threshold of the NMS process. 
In Table~\ref{tab:ablation-nms}, the $\text{IoU}^{h}$ and $\text{IoU}^{o}$ indicate using the human or object bounding boxes between two HOI instances to calculate an IoU, which is similar to the object detection task. 
From the results of $\text{IoU}^{h}$ and $\text{IoU}^{o}$, when using only human or object bounding boxes to filter overlapping HOI instances, the human bounding box can achieve better results.
By using the combined IoU, $\text{IoU}^{h}\cdot \text{IoU}^{o}$ to represent the degree of overlap of two HOI instances, 
the best result can be obtained by setting an IoU threshold of $\delta=0.5$.

\begin{figure*}
  \centering
  \begin{subfigure}{0.18\linewidth}
    \centering
    \includegraphics[height=3.3cm]{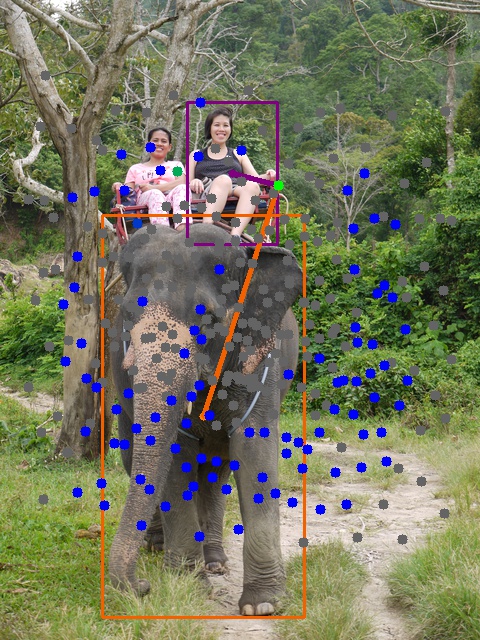}
    \caption{ride, elephant}
    \label{fig:short-a}
  \end{subfigure}
  \begin{subfigure}{0.18\linewidth}
    \centering
    \includegraphics[height=3.3cm]{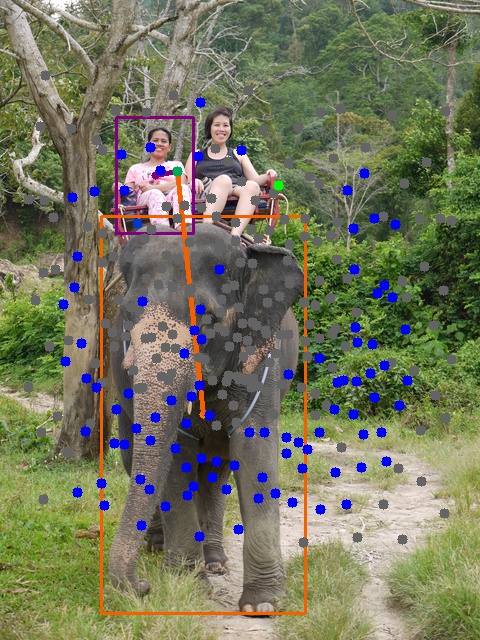}
    \caption{ride, elephant}
    \label{fig:short-b}
  \end{subfigure}
  \begin{subfigure}{0.18\linewidth}
    \centering
    \includegraphics[height=3.3cm]{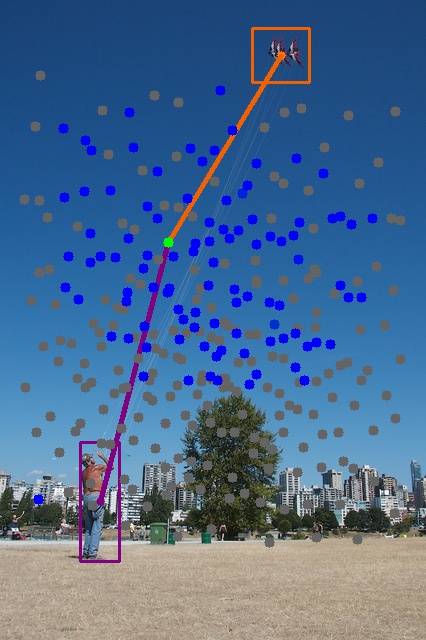}
    \caption{fly, kite}
    \label{fig:short-c}
  \end{subfigure}
  \begin{subfigure}{0.18\linewidth}
    \centering
    \includegraphics[height=3.3cm]{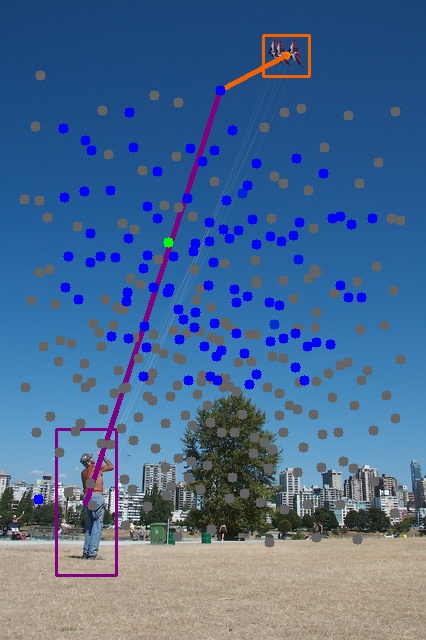}
    \caption{fly, kite}
    \label{fig:short-d}
  \end{subfigure}
  \begin{subfigure}{0.18\linewidth}
    \centering
    \includegraphics[height=3.3cm]{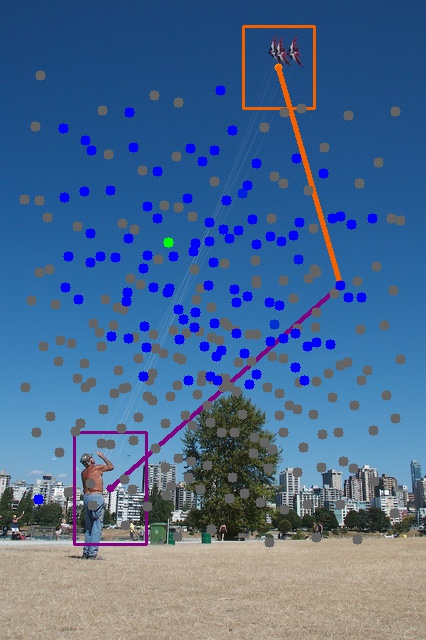}
    \caption{fly, kite}
    \label{fig:short-e}
  \end{subfigure}
  \caption{The flexibility of the anchors. The blue and gray points represent the selected anchors with object class scores of the top 100 and the unselected anchors, and green points represent the anchors with the highest action class scores for each detected HOI instance.}
  \label{fig:anchor-flex}
\end{figure*}
\begin{figure*}
  \centering
  \begin{subfigure}{0.27\linewidth}
    \centering
    \includegraphics[height=2.8cm]{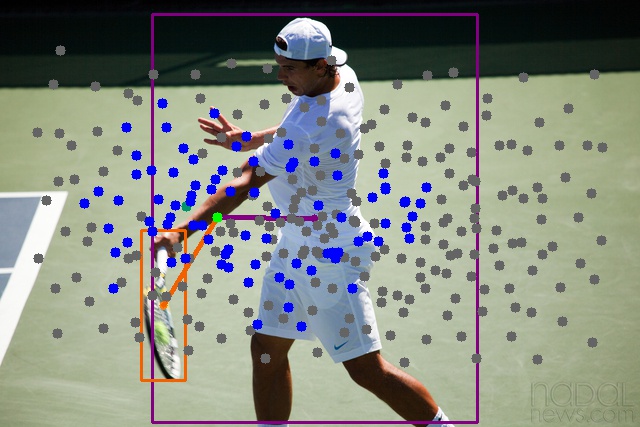}
    \caption{}
    \label{fig:short2-a}
  \end{subfigure}
  \begin{subfigure}{0.27\linewidth}
    \centering
    \includegraphics[height=2.8cm]{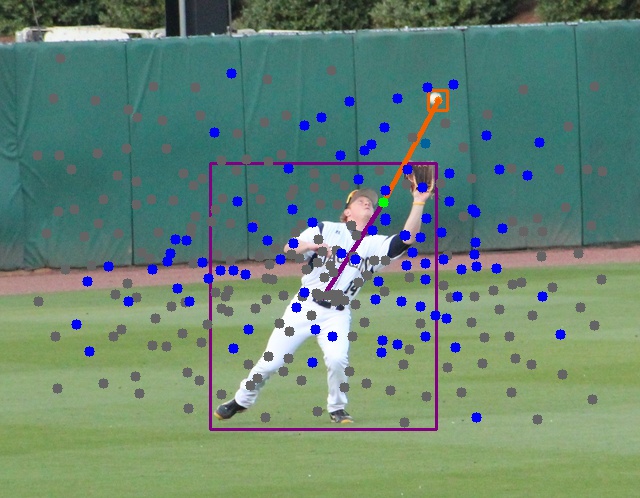}
    \caption{}
    \label{fig:short2-b}
  \end{subfigure}
  \begin{subfigure}{0.27\linewidth}
    \centering
    \includegraphics[height=2.8cm]{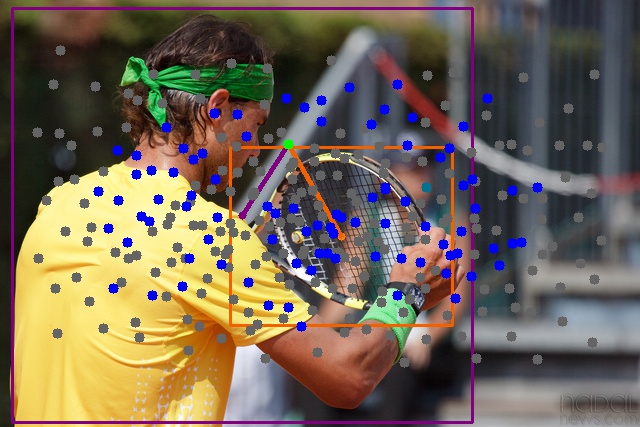}
    \caption{}
    \label{fig:short2-c}
  \end{subfigure}
  \begin{subfigure}{0.15\linewidth}
    \centering
    \includegraphics[height=2.8cm]{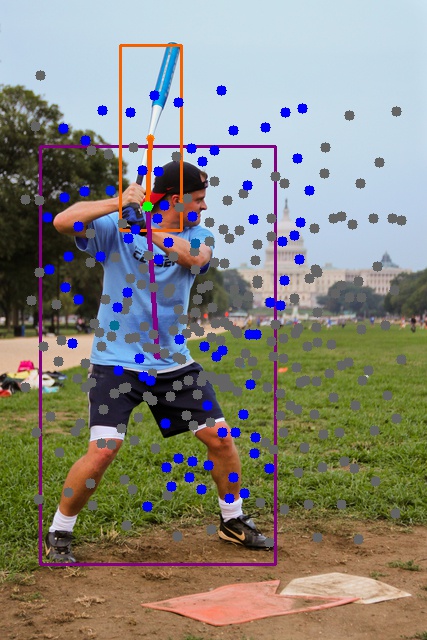}
    \caption{}
    \label{fig:short2-d}
  \end{subfigure}
  \caption{The distribution of the anchors with top 100 object class scores.}
  \label{fig:anchor-top100}
\end{figure*}
\subsection{Contribution at Different Spatial Scales}
The multi-scale architecture of QAHOI should be more advantageous for the detection of small targets. 
In order to investigate the contribution of QAHOI to the detection capability of targets at different spatial scales and compare it with the state-of-the-art transformer-based method, the same as the evaluation method of QPIC, we evaluate both of different center distances and larger areas of HOI instances in different scales.
The results are shown in Figure~\ref{fig:spatial_area} and \ref{fig:spatial_distance}, the ground-truth HOI instances in the test set of HICO-DET is divided into 10 bins, and we also select the bins with more than 1,000 instances to display the AP results.

While the human and object areas are small, it is difficult to extract the features of the area containing interactive information with a low-resolution feature map.
In Figure~\ref{fig:spatial_area}, QAHOI with transformer-based backbones outperform QPIC with ResNet-101 on the detection of small HOI instances in the first three bins. 
QAHOI with ResNet-50, which is fine-tuned from deformable DETR, outperforms the QPIC with ResNet-101 in the first bin and obtains comparable results on the second and third bins.
Besides, QAHOI with Swin-Large and Swin-Base can also perform well on large instances.

The shorter the distance between the human and object of an HOI instance, the harder it is to distinguish features from each other.
In Figure~\ref{fig:spatial_distance}, while the distance between the center of the human and object bounding box is less than 0.3$\times$ image size, QAHOI performs better than QPIC. 
Although the accuracy of both QPIC and QAHOI decreases as the distance between the person and the object increases, a better backbone can alleviate this problem.

\subsection{Qualitative Analysis}
\noindent{\bf The flexibility of Query-Based anchors.}\space The query-based anchors are able to extract features from multi-scale feature maps, which allows anchors to detect HOI instances regardless of their location.
As shown in Figure~\ref{fig:short-a} and \ref{fig:short-b}, two women are riding the elephant, different from the method based on interaction points,
the anchor that detects the action with the highest confidence can be far from the center of the human-object pair but close to the human.
The Figure~\ref{fig:short-c}, \ref{fig:short-d} and \ref{fig:short-e} display the anchors' results with the top 3 action class scores in order. 
In this scene, the human and the object are far from each other, in \ref{fig:short-c} the top 1 anchor in the middle of the human-object pair detects the human well but does not capture the object well,
in \ref{fig:short-d} the top 2 anchor is close to the object but far from the human, and it detects the object well but does not capture the human well,
in \ref{fig:short-e} the anchor is far from both the human and object, and it can locate the human-object pair but does not well detect each of them.
Figure~\ref{fig:short2-a},\ref{fig:short2-b},\ref{fig:short2-c} and \ref{fig:short2-d} further illustrate the distribution of the top 100 anchors for the images with a single HOI instance.
The anchors with high object scores are these located close to the center of the human and object,
however, the anchors with high confidence of the action class are not limited to the center of the human and object, such as the man holding the tennis racket in \ref{fig:short2-a} and \ref{fig:short2-c}.
The quantitative results above show that the query-based anchor is a powerful representation for HOI instances which is flexible in localization.

\section{Conclusion and Future Work}
In this paper, we propose a transformer-based one-stage method for HOI detection,
which leverages a hierarchical backbone and transformer encoder to extract the multi-scale semantic feature,
a transformer decoder to decode the HOI embeddings and an interaction head to predict the HOI instances.
The transformer decoder and the interaction head leverage the query-based anchors to decode the HOI embeddings and predict the HOI instances.
Transformer-based backbones with the attention mechanism show a great advance for HOI detection, and the query-based anchors are also flexible in detecting the HOI instances.

Because our method has a multi-scale architecture and leverages the anchors to detect HOI instances like object detection,
there are several improvements that can be added,
such as the Feature Pyramid Networks (FPN) \cite{lin2017feature} can be added to enhance the multi-scale features,
and the anchors' predictions can be used as the HOI proposals and refining the predictions in a two-stage manner like the two-stage deformable DETR.
Furthermore, without training a detector, our method can train a large model from scratch and achieve a state-of-the-art result.
We hope our method based on query-based anchors can be further developed with the techniques that are used in modern object detectors
and used as a strong baseline for the HOI detection task in future work.

\clearpage
{\small
\bibliographystyle{ieee_fullname}
\bibliography{egbib}

\begin{thebibliography}{10}\itemsep=-1pt

\bibitem{bansal2020detecting}
Ankan Bansal, Sai~Saketh Rambhatla, Abhinav Shrivastava, and Rama Chellappa.
\newblock Detecting human-object interactions via functional generalization.
\newblock In {\em AAAI}, 2020.

\bibitem{carion2020end}
Nicolas Carion, Francisco Massa, Gabriel Synnaeve, Nicolas Usunier, Alexander
  Kirillov, and Sergey Zagoruyko.
\newblock End-to-end object detection with transformers.
\newblock In {\em ECCV}, 2020.

\bibitem{chao2018learning}
Yu-Wei Chao, Yunfan Liu, Xieyang Liu, Huayi Zeng, and Jia Deng.
\newblock Learning to detect human-object interactions.
\newblock In {\em WACV}, 2018.

\bibitem{chen2021reformulating}
Mingfei Chen, Yue Liao, Si Liu, Zhiyuan Chen, Fei Wang, and Chen Qian.
\newblock Reformulating hoi detection as adaptive set prediction.
\newblock In {\em CVPR}, 2021.

\bibitem{deng2009imagenet}
Jia Deng, Wei Dong, Richard Socher, Li-Jia Li, Kai Li, and Li Fei-Fei.
\newblock Imagenet: A large-scale hierarchical image database.
\newblock In {\em CVPR}, 2009.

\bibitem{dosovitskiy2021an}
Alexey Dosovitskiy, Lucas Beyer, Alexander Kolesnikov, Dirk Weissenborn,
  Xiaohua Zhai, Thomas Unterthiner, Mostafa Dehghani, Matthias Minderer, Georg
  Heigold, Sylvain Gelly, Jakob Uszkoreit, and Neil Houlsby.
\newblock An image is worth 16x16 words: Transformers for image recognition at
  scale.
\newblock In {\em ICLR}, 2021.

\bibitem{gao2020drg}
Chen Gao, Jiarui Xu, Yuliang Zou, and Jia-Bin Huang.
\newblock {DRG}: Dual relation graph for human-object interaction detection.
\newblock In {\em ECCV}, 2020.

\bibitem{gao2018ican}
Chen Gao, Yuliang Zou, and Jia-Bin Huang.
\newblock {iCAN}: Instance-centric attention network for human-object
  interaction detection.
\newblock In {\em BMVC}, 2018.

\bibitem{gkioxari2018detecting}
Georgia Gkioxari, Ross Girshick, Piotr Doll{\'a}r, and Kaiming He.
\newblock Detecting and recognizing human-object interactions.
\newblock In {\em CVPR}, 2018.

\bibitem{gupta2019no}
Tanmay Gupta, Alexander Schwing, and Derek Hoiem.
\newblock No-frills human-object interaction detection: Factorization, layout
  encodings, and training techniques.
\newblock In {\em ICCV}, 2019.

\bibitem{he2016deep}
Kaiming He, Xiangyu Zhang, Shaoqing Ren, and Jian Sun.
\newblock Deep residual learning for image recognition.
\newblock In {\em CVPR}, 2016.

\bibitem{hou2020visual}
Zhi Hou, Xiaojiang Peng, Yu Qiao, and Dacheng Tao.
\newblock Visual compositional learning for human-object interaction detection.
\newblock In {\em ECCV}, 2020.

\bibitem{kim2020uniondet}
Bumsoo Kim, Taeho Choi, Jaewoo Kang, and Hyunwoo~J Kim.
\newblock {UnionDet}: Union-level detector towards real-time human-object
  interaction detection.
\newblock In {\em ECCV}, 2020.

\bibitem{kim2021hotr}
Bumsoo Kim, Junhyun Lee, Jaewoo Kang, Eun-Sol Kim, and Hyunwoo~J Kim.
\newblock {HOTR}: End-to-end human-object interaction detection with
  transformers.
\newblock In {\em CVPR}, 2021.

\bibitem{kuhn1955hungarian}
Harold~W Kuhn.
\newblock The hungarian method for the assignment problem.
\newblock {\em Naval Res. Logist. Quart}, pages 83--97, 1955.

\bibitem{li2019transferable}
Yong-Lu Li, Siyuan Zhou, Xijie Huang, Liang Xu, Ze Ma, Hao-Shu Fang, Yanfeng
  Wang, and Cewu Lu.
\newblock Transferable interactiveness knowledge for human-object interaction
  detection.
\newblock In {\em CVPR}, 2019.

\bibitem{liao2020ppdm}
Yue Liao, Si Liu, Fei Wang, Yanjie Chen, Chen Qian, and Jiashi Feng.
\newblock {PPDM}: Parallel point detection and matching for real-time
  human-object interaction detection.
\newblock In {\em CVPR}, 2020.

\bibitem{lin2017feature}
Tsung-Yi Lin, Piotr Doll{\'a}r, Ross Girshick, Kaiming He, Bharath Hariharan,
  and Serge Belongie.
\newblock Feature pyramid networks for object detection.
\newblock In {\em CVPR}, 2017.

\bibitem{lin2017focal}
Tsung-Yi Lin, Priya Goyal, Ross Girshick, Kaiming He, and Piotr Doll{\'a}r.
\newblock Focal loss for dense object detection.
\newblock In {\em ICCV}, 2017.

\bibitem{lin2014microsoft}
Tsung-Yi Lin, Michael Maire, Serge Belongie, James Hays, Pietro Perona, Deva
  Ramanan, Piotr Doll{\'a}r, and C~Lawrence Zitnick.
\newblock {Microsoft COCO}: Common objects in context.
\newblock In {\em ECCV}, 2014.

\bibitem{liu2021swin}
Ze Liu, Yutong Lin, Yue Cao, Han Hu, Yixuan Wei, Zheng Zhang, Stephen Lin, and
  Baining Guo.
\newblock Swin transformer: Hierarchical vision transformer using shifted
  windows.
\newblock In {\em ICCV}, 2021.

\bibitem{loshchilov2018decoupled}
Ilya Loshchilov and Frank Hutter.
\newblock Decoupled weight decay regularization.
\newblock In {\em ICLR}, 2018.

\bibitem{newell2016stacked}
Alejandro Newell, Kaiyu Yang, and Jia Deng.
\newblock Stacked hourglass networks for human pose estimation.
\newblock In {\em ECCV}, 2016.

\bibitem{qi2018learning}
Siyuan Qi, Wenguan Wang, Baoxiong Jia, Jianbing Shen, and Song-Chun Zhu.
\newblock Learning human-object interactions by graph parsing neural networks.
\newblock In {\em ECCV}, 2018.

\bibitem{ren2016faster}
Shaoqing Ren, Kaiming He, Ross Girshick, and Jian Sun.
\newblock {Faster R-CNN}: Towards real-time object detection with region
  proposal networks.
\newblock In {\em IEEE TPAMI}, 2016.

\bibitem{tamura2021qpic}
Masato Tamura, Hiroki Ohashi, and Tomoaki Yoshinaga.
\newblock {QPIC}: Query-based pairwise human-object interaction detection with
  image-wide contextual information.
\newblock In {\em CVPR}, 2021.

\bibitem{ulutan2020vsgnet}
Oytun Ulutan, ASM Iftekhar, and Bangalore~S Manjunath.
\newblock {VSGNet}: Spatial attention network for detecting human object
  interactions using graph convolutions.
\newblock In {\em CVPR}, 2020.

\bibitem{wan2019pose}
Bo Wan, Desen Zhou, Yongfei Liu, Rongjie Li, and Xuming He.
\newblock Pose-aware multi-level feature network for human object interaction
  detection.
\newblock In {\em ICCV}, 2019.

\bibitem{wang2020max}
Huiyu Wang, Yukun Zhu, Hartwig Adam, Alan Yuille, and Liang-Chieh Chen.
\newblock {MaX-DeepLab}: End-to-end panoptic segmentation with mask
  transformers.
\newblock In {\em CVPR}, 2021.

\bibitem{wang2020learning}
Tiancai Wang, Tong Yang, Martin Danelljan, Fahad~Shahbaz Khan, Xiangyu Zhang,
  and Jian Sun.
\newblock Learning human-object interaction detection using interaction points.
\newblock In {\em CVPR}, 2020.

\bibitem{waswani2017attention}
A Waswani, N Shazeer, N Parmar, J Uszkoreit, L Jones, AN Gomez, L Kaiser, and I
  Polosukhin.
\newblock Attention is all you need.
\newblock In {\em NeurIPS}, 2017.

\bibitem{yu2018deep}
Fisher Yu, Dequan Wang, Evan Shelhamer, and Trevor Darrell.
\newblock Deep layer aggregation.
\newblock In {\em CVPR}, 2018.

\bibitem{zhang2020spatio}
Frederic~Z Zhang, Dylan Campbell, and Stephen Gould.
\newblock Spatially conditioned graphs for detecting human-object interactions.
\newblock In {\em ICCV}, 2021.

\bibitem{zhong2021polysemy}
Xubin Zhong, Changxing Ding, Xian Qu, and Dacheng Tao.
\newblock Polysemy deciphering network for robust human-object interaction
  detection.
\newblock In {\em IJCV}, 2021.

\bibitem{zhong2021glance}
Xubin Zhong, Xian Qu, Changxing Ding, and Dacheng Tao.
\newblock {Glance and Gaze}: Inferring action-aware points for one-stage
  human-object interaction detection.
\newblock In {\em CVPR}, 2021.

\bibitem{zhou2019relation}
Penghao Zhou and Mingmin Chi.
\newblock Relation parsing neural network for human-object interaction
  detection.
\newblock In {\em ICCV}, 2019.

\bibitem{zhu2020deformable}
Xizhou Zhu, Weijie Su, Lewei Lu, Bin Li, Xiaogang Wang, and Jifeng Dai.
\newblock {Deformable DETR}: Deformable transformers for end-to-end object
  detection.
\newblock In {\em ICLR}, 2020.

\bibitem{zou2021end}
Cheng Zou, Bohan Wang, Yue Hu, Junqi Liu, Qian Wu, Yu Zhao, Boxun Li, Chenguang
  Zhang, Chi Zhang, Yichen Wei, et~al.
\newblock End-to-end human object interaction detection with hoi transformer.
\newblock In {\em CVPR}, 2021.

\end{thebibliography}
}

\end{document}